\documentclass{article} 
\usepackage{iclr2021_conference,times}

\usepackage{amsmath,amsfonts,bm}

\def\eqref#1{equation~\ref{#1}}

\def\1{\bm{1}}

\DeclareMathAlphabet{\mathsfit}{\encodingdefault}{\sfdefault}{m}{sl}
\SetMathAlphabet{\mathsfit}{bold}{\encodingdefault}{\sfdefault}{bx}{n}

\usepackage{url}
\usepackage{hyperref}

\usepackage{mathtools}
\usepackage{multirow,tabularx}
\usepackage{amsmath}
\usepackage{subcaption}
\usepackage[final]{graphicx}

\usepackage{wrapfig}

\newcommand{\Mat}[1]{\mathbf{#1}}

\newcommand{\Space}[1]{\mathbb{#1}}
\newcommand{\Set}[1]{\mathcal{#1}}

\newcommand{\ie}{\emph{i.e., }}
\newcommand{\eg}{\emph{e.g., }}

\newcommand{\wrt}{\emph{w.r.t. }}
\newcommand{\cf}{\emph{cf. }}

\title{A-FMI: Learning Attributions From Deep\\Networks via Feature Map Importance}

\iclrfinalcopy

\author{An Zhang$^1$, Xiang Wang$^1$, Chengfang Fang$^2$, Jie Shi$^2$, Tat-seng Chua$^1$, Zehua Chen$^2$\\
    $^1$National University of Singapore, $^2$Huawei Singapore Research Center \\
    \texttt{anzhang@u.nus.edu,xiangwang@u.nus.edu,fang.chengfang@huawei.com,}\\
    \texttt{shi.jie1@huawei.com,dcscts@nus.edu.sg,stachenz@nus.edu.sg}
}

\begin{document}

\maketitle

\begin{abstract}
Gradient-based attribution methods can aid in the understanding of convolutional neural networks (CNNs).
However, the redundancy of attribution features and the gradient saturation problem, which weaken the ability to identify significant features and cause an explanation focus shift, are challenges that attribution methods still face.
In this work, we propose: 1) an essential characteristic, Strong Relevance, when selecting attribution features;
2) a new concept, feature map importance (FMI), to refine the contribution of each feature map, which is faithful to the CNN model;
and 3) a novel attribution method via FMI, termed \textbf{A-FMI}, to address the gradient saturation problem, which couples the target image with a reference image, and assigns the FMI to the ``difference-from-reference'' at the granularity of feature map.
Through visual inspections and qualitative evaluations on the ImageNet dataset, we show the compelling advantages of A-FMI on its faithfulness, insensitivity to the choice of reference, class discriminability, and superior explanation performance compared with popular attribution methods across varying CNN architectures.
\end{abstract}

\section{Introduction}


As the understanding of neural networks is of crucial importance to engender user trust, interpreting network behavior has attracted increasing attention.
To this end, attribution methods~\citep{attribution_method} have demonstrated the remarkable ability in attributing the prediction of a given network, typically CNN, to its input.
Regardless of various designs, a common axiom called completeness~\citep{IG} or local-faithfulness~\citep{Grad-CAM} for most existing attribution methods can be summarized as:
$\sum$\emph{attribution feature}$\times$\emph{attribution score}$\approx$\emph{network prediction}.
Simply stated, given an input image, attribution method determines the attribution score (also well-known as relevance~\citep{LRP} or contribution~\citep{LIME}) of each attribution feature (\eg pixel, segment), in order to approximate the CNN's prediction of interest.
By redistributing such attributions to the input image, we can produce a saliency map to highlight the most important regions for predicting the class of interest. 




In scrutinizing the axiom, we argue that a key challenge lies in the selection of attribution features, which should satisfy one essential characteristic --- Strong Relevance (a statistic concept in feature selection~\citep{Relevant}). Formally, a strong relevant feature is defined as a feature which has information that is both pertinent to the class prediction and can barely be derived from other features;
in contrast, the redundant feature is highly correlated and can be represented by other features.
Although the redundancy in attribution features does not affect the aprroximation of the prediction, it negatively influences the learning of the attribution scores.
The key reason is that the entanglement among these features weakens the ability to identify significant features, and small changes of redundant features can swing the value of attribution scores widely.

\begin{figure}[t]
 \centering
 \includegraphics[width=\textwidth]{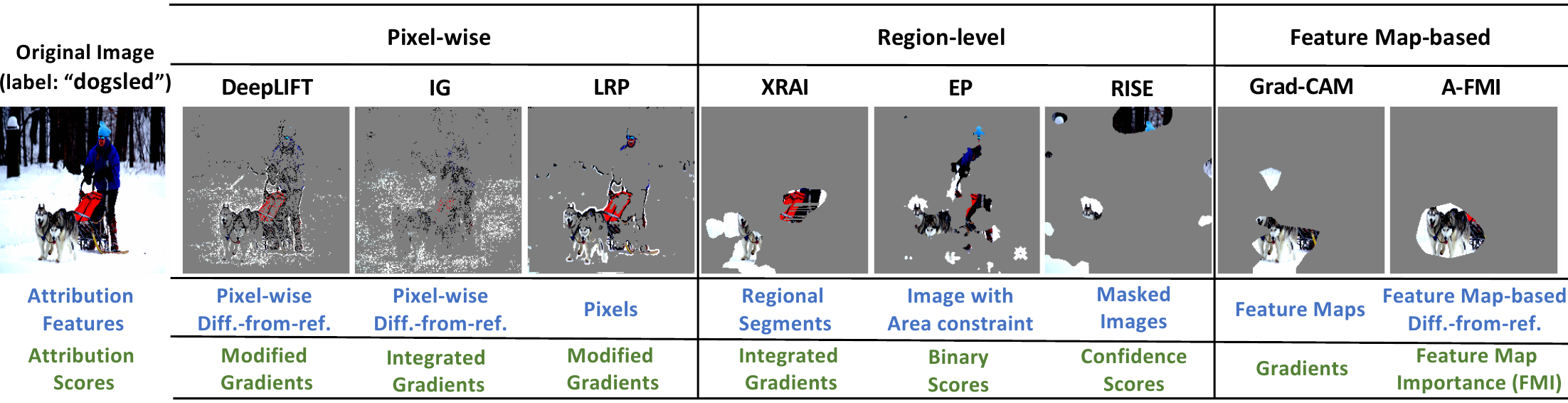}
 \vspace{-13px}
 \caption{Visual comparison of present attribution methods and the proposed A-FMI at 10\% of important pixels insertion for the prediction of class ``dogsled''. Pixels with significant attributions are remained, while the rest is removed. A black reference is used for DeepLIFT, IG and A-FMI.}
 \vspace{-15px}
 \label{fig:introduction}
\end{figure}

Having realized the vital role of attribution feature selection, we categorize the prior works into three levels of granularity: \textbf{pixels}~\citep{gradients,Deconvolution,GuidedBackpropogation,LRP,gradients*input,SmoothGrad,DeepLIFT,IG,SHAP}, \textbf{regions}~\citep{LIME,PredictionDifferenceAnalysis,MP,real_time_saliency, RISE,EP, XRAI}, and \textbf{feature maps}~\citep{CAM,Grad-CAM,Grad-CAM++}. However, each type of methods suffers from some limitations:
\begin{itemize}
    \itemsep0em
    \item One research line exploits single pixels as attribution features and learns the gradients, modified gradients or integrated gradients of the target class as the attribution scores.
    Despite great success, one limitation is that it violates the Strong Relevance characteristic of attribution features, since pixels are strongly related to thier surrounding pixels.
    Such redundancy easily results in suboptimal and fragile attributions, being akin to the edge detector~\citep{SanityCheck} and vulnerable to small perturbations~\citep{fragile}.

    \item Region-level methods measure how sensitive the prediction is to the masked image or perturbations of regional segments.
    However, they hardly satisfy the completeness principle, \ie adding the attributions of all regions up may not guarantee a match with the prediction. 
    Moreover, built upon the perturbation or masking mechanism, they are time-consuming and heavily depend on the quality of segmentation.
    
    
    \item Decomposing the target prediction to the feature maps in the last convolutional layer via gradient learning has been studied. Compared with pixel-wise attribution methods that easily result in focus shift, feature map-based methods highlight the compact regions of interest (see Figure~\ref{fig:introduction}). We attribute such success to the Strong Relevance characteristic of feature maps, wherein only subtle correlations exist as shown in~\cite{DeepVisualization, FeatureMapRelevant}. However, previous methods inherit gradient saturation problem, which underestimates the contributions and further leading untargeted explanations (see Figure~\ref{fig:framework}).

\end{itemize}


In this work, feature map referece is introduced as a solution to the gradient saturation problems.
More specifically, we implement a baseline input (\eg a black image), run a forward pass to get feature map references, and measure the differences with the original feature maps.
To refine the contributions for feature map-based difference-from-references, we devise \textbf{feature map importance} (FMI), which is a set of modified gradients estimated by the Taylor series.
FMI allows significant feature maps to propagate signals even in situations where their gradients are zero.
Moreover, FMI has notable properties (evidence in Section~\ref{sec:explanation-faithfulness}): 1) strong representation ability, embeding class-specific information which could be directly used to classify, 2) local faithfulness, \ie in the vicinity of the input image, FMI is locally accurate to the CNN.

We further propose a new \textbf{Attribution Method via FMI} (A-FMI), which is equipped with difference-from-reference and FMI to interpret the CNN models.
To validate the effectiveness of A-FMI, we perform thorough visual inspections and qualitative evaluations for VGG19 and ResNet50 on the ImageNet dataset.
The empirical results demonstrate that A-FMI consistently produces better results than popular attribution methods and is conceptually advantageous in that:
1) compared with pixel-wise methods, it uses attribution features (\ie feature maps) with Strong Relevance characteristic, thus able to output compact regions of interest; moreover, distinct from other attribution methods who are sensitive to the selection of reference, A-FMI is more robust (evidence in Section~\ref{sec:insensitivity-reference});
2) compared with region-level methods, it satisfies the local faithfulness and is much more efficient;
and 3) compared with feature map-based methods, it inherits the advantage of class discriminability (evidence in Section~\ref{sec:class-discriminability}) and solves the gradient saturation problem, producing more refined and targeted explanations. In summary, this work makes the following contributions:
\begin{itemize}
    \itemsep0em
    \item We emphasize the important role of the selection of attribution features, and compare attribution features into three levels of granularity: pixels, regions, and feature maps \wrt the desirable Strong Relevance characteristic.
    \item To the best of our knowledge, we are the first to introduce the feature map reference and FMI into the attribution methods, to solve the gradient saturation problem.
    \item We conduct extensive experiments, demonstrating the effectiveness and consistency of A-FMI in interpreting VGG-based and ResNet-based architectures on the ImageNet dataset.
\end{itemize}

\section{Related Work}
Pixel-wise attribution methods largely leverage the backpropagation way to redistribute the prediction through the whole CNN model to single pixels.
For example, Gradient \citep{gradients} and Input*Gradient \citep{gradients*input} use gradient of the prediction \wrt each pixel as attribution scores;
DeconvNet~\citep{Deconvolution} and Guided Backpropogation~\citep{GuidedBackpropogation} employ well-designed operations on the gradients of nonlinear activation functions. 
However, the gradient saturation problem is inherent in the backpropagation way, which easily results in the vanishing gradients and underestimating importance of pixels.
To solve this problem, DeepLIFT~\citep{DeepLIFT} employs a baseline (reference) image to calculate the modified partial derivatives of the difference-from-reference as the importance of pixels;
meanwhile, Integrated Gradients (IG)~\citep{IG} aggregates the gradients by gradually varying the input from the baseline to the original image.
Despite great success, the significant pixels highlighted by pixel-wise attribution methods are easily spread out --- that is, the focus of the explanation model shifts into irrelevant edges, objects, or even background (see Figure~\ref{fig:introduction}, pixel-wise attribution methods select top 10\% important pixels including snow and person for label dogsled).


Region-level attribution methods combine single pixels into patches or regions, and mainly apply the perturbation mechanism to directly evaluate the marginal effect of each region by masking or replacing it.
For example, LIME~\citep{LIME} approximates the CNN function by a sparse linear model between the patches and prediction, which is learned on perturbations of patches.
Prediction Difference Analysis~\citep{PredictionDifferenceAnalysis} replaces patches with a sample from other images and obtain the contribution of each pixel by averaging the importance of patches containing the pixel.
RISE~\citep{RISE} randomly occludes an image and weights the changes in the confidence score of CNN.
Extremal Perturbations (EP)~\citep{EP} optimizes a spatial perturbation mask with a fixed area that maximally affects the CNN's output.
XRAI~\citep{XRAI} incorporates the idea of gradients with the region-level features, coalescing smaller regions into larger segments based on the maximum gain of IG per region.
These methods suffer from high computational complexity and heavily rely on the quality of segmentation.


Feature map-based attribution methods use feature maps in the last convolutional layer to produce saliency maps.
Wherein, feature maps capture high-level semantic patterns and spatial information.
CAM~\citep{CAM} replaces the fully-connected layers with a global average pooling layer, and produces a coarse-grained saliency map via a weighted sum of feature maps.
Furthermore, Grad-CAM~\citep{Grad-CAM} uses the average gradients \wrt feature maps as the attribution scores.
Though using gradients is beneficial, Grad-CAM inherits the gradient saturation problem, which is unexplored in the existing feature map-based methods.
In this work, the proposed method, A-FMI, falls within this family and provides a potential solution to the saturation problem.

\section{Methodology}



\noindent\textbf{Problem Formulation.}
Let $F$ be the original CNN model that classifies an input image $\Mat{x}$ into one of $C$ classes.
The local explanation model designs to interpret $F(\Mat{x})$ consisting of two components: attribution features and attribution scores. 
We utilize feature maps $\Set{A}$ in the last convolutional layer as the attribution features, obtained by feeding $\Mat{x}$ into the CNN.
Our goal is to learn the local explanation model (or instance-wise saliency map) $\Mat{S}^{c}(\Set{A})$ for a class of interest $c$ that ensures local faithfulness:  $\sum_{i,j}S^{c}(\Set{A})_{i,j}\approx F^{c}(\Mat{x})$. 
In other words, the sum of entries in the saliency map should approximate the prediction of interest. 

\noindent\textbf{Gradient Saturation Problem.}\label{sec:saturation-problem}
Exploiting the gradients of attribution features \wrt the target class as attribution scores is a prevalent technique in the existing attribution methods.
However, the problem of gradient saturation is inherent in the gradient learning process, which easily leads to suboptimal attribution scores.
The key reason is that, during the backward pass, vanishing gradients may occur due to nonlinear activation functions (\eg sigmoid, tanh, and even ReLU).
Taking a two-layer ReLU network $F(\alpha)=\text{ReLU}(-\text{ReLU}(-\alpha+1)+2)$ as an example, the gradient of $F(\alpha)$ at $\alpha=2$ is $0$, intuitively indicating its trivial contribution.
However, when changing $\alpha$ from $0$ to $2$, the network output changes from $F(\alpha=0)=1$ to $F(\alpha=2)=2$, which suggests that $\alpha$ is significant for prediction.
Clearly, such vanishing gradients easily mislead the learning of attribution scores.
Hence, It is of crucial importance to solve the gradient saturation problem.


\begin{figure}[t]
    \centering
    \includegraphics[width=\linewidth]{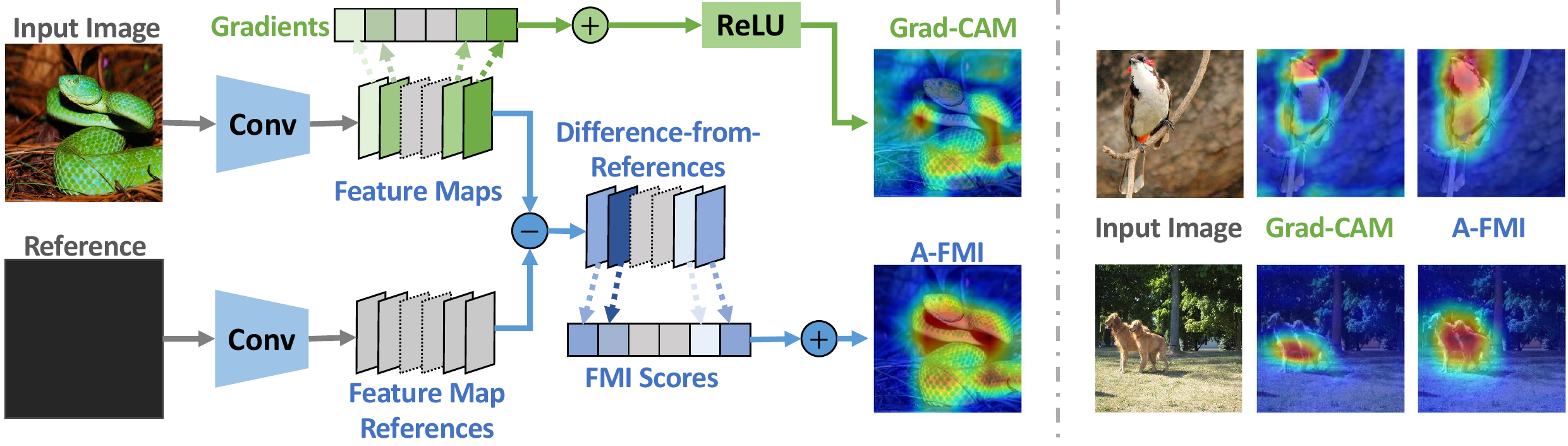}
    \vspace{-13px}
    \caption{Overview (left) and Visual Comparisons (right) between Grad-CAM and A-FMI.} 
    \vspace{-15px}
    \label{fig:framework}
\end{figure}

\noindent\textbf{Mathematical Motivation.}
For clarification of presentation, let us first consider a network that only consists of $L$ fully connected layers as the classifier being interpreted, while first leaving the feature maps untouched.
Let $\Mat{x}^{(0)}\in\Space{R}^{d_{0}}$ denote the input vector associated with $d_{0}$-dimensional features, and $\Mat{y}^{(L)}\in\Space{R}^{C}$ be the prediction over $C$ classes in the last layer (before softmax).
In a forward pass, the output of each layer $\Mat{x}^{(l)}$ is modeled as:
\begin{equation}
    \Mat{x}^{(l)}=\sigma{(\Mat{y}^{(l)})},\quad\Mat{y}^{(l)}=\Mat{W}^{(l)}\Mat{x}^{(l-1)} + \Mat{b}^{(l)}
\end{equation}
where $\Mat{W}^{(l)}\in\Space{R}^{d_{l}\times d_{l-1}}$ and $\Mat{b}^{(l)}\in\Space{R}^{d_{l}}$ are the weight matrix and bias terms in the $l$-th layer; $\sigma(\cdot)$ is a nonlinear activation function; and $d_{l}$ is the dimension of the $l$-th layer.
Obviously, the $j$-th neuron in the $l$-th layer is represented using all neurons in the previous layer as: $x_j^{(l)} = \sigma(y_j^{(l)}) = \sigma(\sum_i W_{ji}^{(l)}x_i^{(l-1)}+b_j^{(l)})$.
Hence, to measure the contribution of the $i$-th neuron in ($l$-1)-th layer (\ie $x^{(l-1)}_{i}$) to the $j$-th neuron in the subsequent layer (\ie $x^{(l)}_{j}$), an intuitive solution is to leverage the partial derivative as:
$\frac{\partial x_j^{(l)}}{\partial x_i^{(l-1)}}=\frac{\partial x_j^{(l)}}{\partial y_j^{(l)}} \cdot \frac{\partial y_j^{(l)}}{\partial x_i^{(l-1)}} = \sigma'(y_j^{(l)})\cdot W_{ji}^{(l)}$.
However, when $\sigma(y_j^{(l)})$ is flatten, $\sigma'(y_j^{(l)})$ approaches zero which may underestimate the contribution. To avoid unreasonable gradient vanishing, we employ the first order Taylor series to better estimate $\sigma'(y_j^{(l)})$ using a reference point $\bar{y}_{j}^{(l)}$, as: 
\begin{equation}
    \hat{\sigma'}(y_j^{(l)}) = \frac{\sigma(y_j^{(l)})-\sigma(\bar{y}_{j}^{(l)})}{y_j^{(l)}-\bar{y}_{j}^{(l)}}
\end{equation}
Here we use two cases to illuminate how the estimation process is of promise to solving the saturation problem:
1) When the activation function is ReLU,
$\hat{\sigma'}(y_j^{(l)}) = \frac{\max\{0,y_j^{(l)}\}-\max\{0,\bar{y}_{j}^{(l)}\}}{y_j^{(l)}-\bar{y}_{j}^{(l)}}$.
It always assigns a value in a range of $(0,1)$ when the signs of $y_j^{(l)}$ and $\bar{y}_{j}^{(l)}$ are different; and it is zero if and only if $\sigma(y_j^{(l)})=\sigma(\bar{y}_{j}^{(l)})$.
2) When the activation function is tanh, we set the reference as 0 for simplicity and have
$\hat{\sigma'}(y_j^{(l)}) = \frac{1}{y_j^{(l)}} \int_0^{y_j^{(l)}}\sigma'(t)dt$,
which is the average gradient of the nonlinearity in $[0,y_j^{(l)}]$.
These estimators show the ability in identifying the significant features by assigning a non-zero value and filtering the irrelevant ones out by outputting a zero value.

Having established the estimator $\hat{\sigma'}(y_j^{(l)})$, we are able to approximate the difference of neuron $x_j^{(l)}$ from its own reference $\bar{x}_{j}^{(l)}$ in the $l$-th layer using all neurons in the ($l$-1)-th layer as follows:
\begin{gather}
    x_j^{(l)}-\bar{x}_{j}^{(l)}=\sum_{i} \hat{\sigma'}(y_j^{(l)}) \cdot w_{ji}^{(l)} \cdot (x_i^{(l-1)}-\bar{x}_{i}^{(l-1)})
\end{gather}
Consequently, we can define the modified gradient of the $l$-layer's output $x_j^{(l)}$ \wrt $x_i^{(l-1)}$ as $\frac{\partial^{\sigma} x_j^{(l)}}{\partial x_i^{(l-1)}} = \sum_{i} \hat{\sigma'}(y_j^{(l)}) \cdot w_{ji}^{(l)}$.
We further apply the chain rule of gradient to estimate the attribution of the original input $x_i^{(0)}$ to the final prediction $y^{(L)}_{c}$ of the last layer (before softmax) as:
\begin{gather}\label{equ:generic-attribution-score}
    \frac{\partial ^{\sigma} y^{(L)}_{c}}{\partial x_i^{(0)}} = \sum_{p \in \Set{P}} (\prod \hat{\sigma'_{p}} \prod W_{p})
\end{gather}
where $\Set{P}$ is the set of paths that connect the $i$-th neuron in the input (\ie $x_i^{(0)}$) with the $c$-th neuron in the last layer (\eg $y^{(L)}_{c}$).
Iteratively, in the last layer, the difference between $y^{(L)}_{c}$ and its reference $\bar{y}^{(L)}_{c}$ can be successfully fetched by a weighted sum of all inputs:
\begin{gather}\label{equ:generic-attribution-objective}
    y^{(L)}_{c} - \bar{y}^{(L)}_{c}=\sum_i \frac{\partial ^{\sigma} y^{(L)}_{c}}{\partial x_i^{(0)}} \cdot (x_i^{(0)}-\bar{x}_{i}^{(0)})
\end{gather}
An image that has no specific information for any class, \ie has a near-zero prediction score, can be a proper reference. This encourages the sum of attributions to approximate the prediction of interest. 
Moreover, unlike previous studies that mostly treat instance-wise baselines as hyper-parameters and need to carefully select them (\eg adversarial examples or blurred images) to improve the explanation quality, we simply fix an identical reference for all images, to which our attribution scores are insensitive.

\noindent\textbf{Attribution Method via FMI (A-FMI).}
Having obtained the potential solution to the gradient saturation problem, we subsume our feature map-based attribution method under this solution, as Figure~\ref{fig:framework} shows.
Distinct from Grad-CAM that only takes the target image as input, we additionally couple it with a reference image (\eg a black image) and feed them into the CNN model.
The CNN model yields the feature maps $\Set{A}=\{\Mat{A}^{k}\}$ with the feature map references $\bar{\Set{A}}=\{\bar{\Mat{A}}^{k}\}$ in the last convolutional layer, and the outputs $\{\Mat{x}^{(L)}\}$ and output references $\{\bar{\Mat{x}}^{(L)}\}$ (before softmax) in the last fully-connected layers.
Thereafter, to obtain the $\text{FMI}_{k}^{c}$ of the $k$-th feature map $\Mat{A}^{k}\in\Set{A}$ to the target prediction $y_{c}^{(L)}$, we simply change the single neuron $x^{(0)}_{i}$ in Equation~\eqref{equ:generic-attribution-score} with the entry of $\Mat{A}^{k}_{ij}$ and average all modified gradients in the feature map $\Mat{A}^{k}$, which is formulated as follows:
\begin{gather}\label{equ:fmi}
    \text{FMI}_{k}^{c} = \frac{1}{N} \sum_{i} \sum_{j} \frac{\partial^{\sigma} y_c^{(L)}}{\partial A^k_{ij}} = \frac{1}{N} \sum_{i} \sum_{j} \sum_{p \in P} (\prod \hat{\sigma'_{p}} \prod W_{p}),
\end{gather}
where $A_{ij}^{k}$ is the entry of $\Mat{A}^{k}$ located in the $i$-th row and $j$-th column; $N$ is the total number of entries in the feature map $\Mat{A}^{k}$; and $\Set{P}$ is the set of paths starting from any neuron in this feature map to the prediction score $y_{c}^{(L)}$.
In essence, $\text{FMI}_{k}^c$ is the modified weight of the fully connected layers in the CNN model, which directly captures the contribution of feature map $\Mat{A}^{k}$ in predicting target class $c$. 

The saliency map $S_{\text{A-FMI}}^c$ is modeled as a weighted linear combination of the difference between the feature maps $\{\Mat{A}^{k}\}$ and feature map references $\{\Mat{\bar{A}}^{k}\}$ with the corresponding $\text{FMI}_{k}^c$ scores as the coefficients.
More formally, the weighted combination can be represented as follows:
\begin{equation}
    \Mat{S}_{\text{A-FMI}}^c = \sum_{k} \text{FMI}_{k}^c \cdot (\Mat{A}^k - \Mat{\bar{A}}^{k})
\end{equation}
Note that feature map importance $\text{FMI}_{k}^c$ differs from the importance weights in Grad-CAM.  

\section{Experiments}
In this section, we conduct extensive experiments to answer the following four research questions:
\begin{itemize}
    \itemsep0em
    \item \textbf{RQ1}: Are the explanations of A-FMI faithful to the CNN model?
    \item \textbf{RQ2}: How do different reference images affect the explanations of A-FMI? 
    \item \textbf{RQ3}: Can A-FMI distinguish between different classes of interest?
    \item \textbf{RQ4}: How does A-FMI perform compared with other popular attribution methods?
\end{itemize}

To answer RQ1, we qualitatively evaluate the faithfulness of A-FMI to the VGG19~\citep{vgg-19} model on the ImageNet~\citep{ImageNet} validation set, as well as a simple CNN model~\citep{pytorch_cnn} on the MNIST~\citep{MNIST} validation set, where only correctly predicted images are considered (\cf Section~\ref{sec:explanation-faithfulness}).
To answer the remaining questions, we compare A-FMI to popular attribution methods:
pixel-wise (Gradient, DeepLIFT, IG, LRP),
region-level (XRAI, EP, RISE), and feature map-based (Grad-CAM).
Without specification, we use a black image as the reference in DeepLIFT, IG, and A-FMI.
Visual inspections on VGG19 and qualitative metrics on both VGG19 and ResNet50~\citep{ResNet} are used to validate the effectiveness of A-FMI in terms of reference reliability (\cf Section~\ref{sec:insensitivity-reference}), class discriminability (\cf Section~\ref{sec:class-discriminability}), and explanation quality (\cf Section~\ref{sec:overall-performance}). 
To ensure the reproducibility of our work, we have uploaded the code of A-FMI, all baselines, and their comparisons in the supplementary material.

\subsection{Faithfulness of Explanation (RQ1)}\label{sec:explanation-faithfulness} 
The faithfulness of an explanation model is its ability to accurately estimate the function learned by the CNN model.
Typically, it can be described on two levels: 1) local faithfulness, justifying whether the explanations corresponds to the CNN predictions in the vicinity of an image instance, and 2) global faithfulness, validating whether the globally important features for a target class are identified.
However, as the evaluation of global faithfulness is seldom performed in existing attribution methods, no measuring framework is available.
Hence, we propose a measuring framework which uses FMI to classify the images to evaluate the faithfulness of A-FMI from both local and global perspectives.

To this end, for a specific image in class $c$, we use the attribution scores of its feature maps $\text{FMI}^c = \{\text{FMI}_1^c, \cdots, \text{FMI}_{K}^c\}$ as the refined representation of the image.
We then average the class-specific attribution scores over all training images as $\overline{\text{FMI}}^c$, which can be viewed as the representation of the prototype in class $c$.
Thereafter, we classify each image in the validation set based on its cosine similarity with the prototype representations $\{\overline{\text{FMI}}^1,\cdots,\overline{\text{FMI}}^C\}$.
If the FMI-based class is identical to the target class $c$, the explanation of an image is faithful to the CNN model.
Impressively, we achieve an average explanation accuracy of \textbf{86.9\%} and \textbf{88.4\%} in MNIST with 10 classes and ImageNet datasets with 1000 classes, respectively.

We present a faithfulness analysis of A-FMI based on the classification accuracy.
First, the instance-wise $\text{FMI}^c$ is well qualified to be a representation of an image, which encodes the information pertinent to the class $c$ from a single image and hence reflects the local faithfulness.
Then, the cosine similarity-based accuracy indicates that images with the same classes tend to form clusters, and the prototype representation $\overline{\text{FMI}}^c$ captures the class-wise patterns.
This further suggests that A-FMI to some extent achieves the global faithfulness.

\begin{figure}[t]
    \centering
    \includegraphics[width=\linewidth]{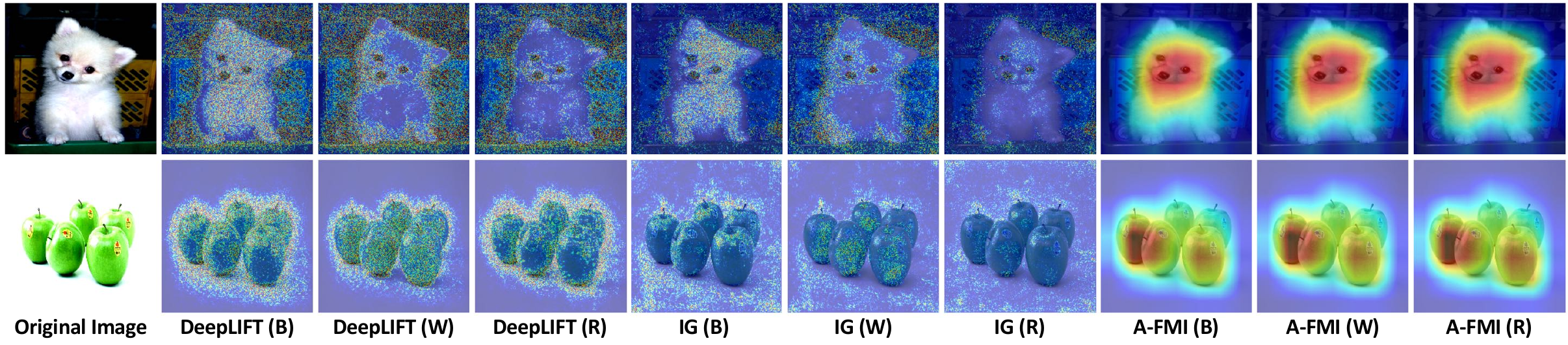}
    \vspace{-10px}
    \caption{\textbf{Reference reliability.} A-FMI is reliable for the different choices of reference, while both DeepLIFT and IG determine the attribution strongly relative to a chosen reference. B, W, and R stand for black, white, and random reference respectively. See more samples in Appendix C.} 
    \vspace{-15px}
    \label{fig:reference}
\end{figure}

\subsection{Reference Reliability (RQ2)}\label{sec:insensitivity-reference}
To engender user trust, a reliable and trustworthy explanation model should be robust to factors that do not contribute to the model prediction.
Hence, we explore how the reference, a factor additionally introduced to solve the saturation problem, affects DeepLIFT, IG, and A-FMI.
Accordingly, we consider the variants of A-FMI, DeepLIFT, and IG that use different references --- a black image, a white image, and an image filled with random pixels to produce saliency maps in Figure~\ref{fig:reference}.

We observe that DeepLIFT and IG shift their focus when using different references. In particular, the explanations of DeepLIFT and IG with white reference tend to focus on the darker pixels and vice versa. For example, DeepLIFT(W) and IG(W) pay more attention to the black background than the white dog (1st row), while DeepLIFT(B) and IG(B) emphasize the white background surrounding the green apples (2nd row). Moreover, the results of DeepLIFT(R) and IG(R) are unsatisfactory when utilizing a random reference. This verifies that the selection of reference significantly affects the explanations of DeepLIFT and IG.
In contrast, the saliency maps of A-FMI are more consistent, suggesting that the proper reference acts as a prior for A-FMI rather than a hyperparameter.
This validates the robustness and reliability of A-FMI without hinging on instance-by-instance solutions~\citep{input_invariance}. 

\subsection{Class Discriminability (RQ3)}\label{sec:class-discriminability}

\begin{figure}[t]
    \centering
    \includegraphics[width=\linewidth]{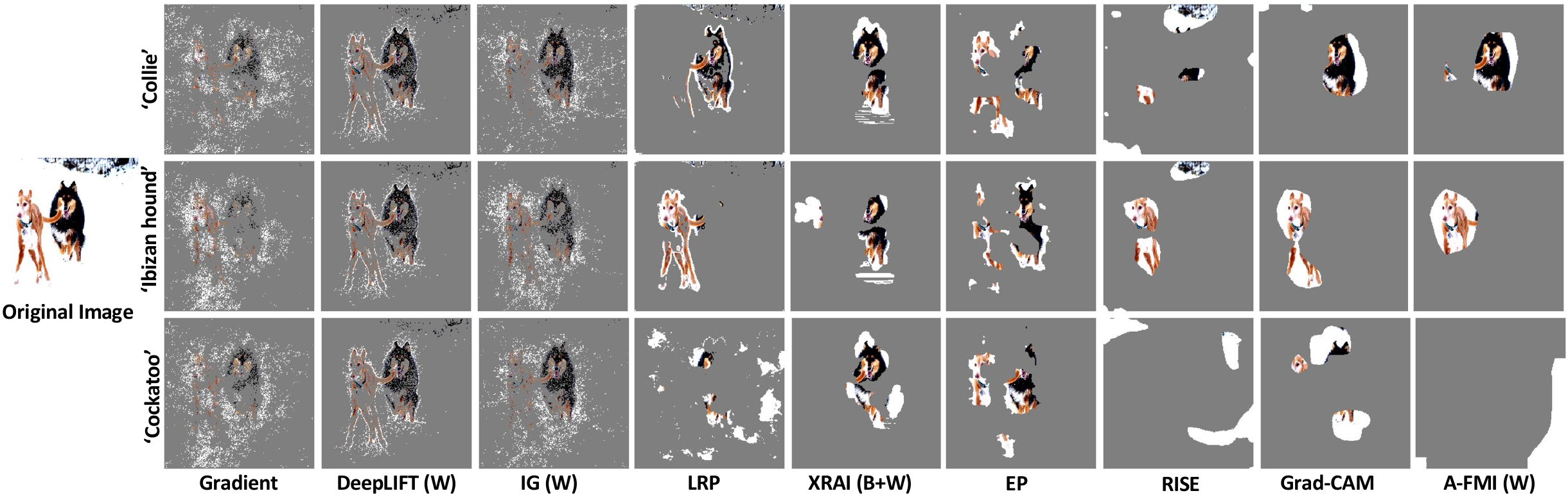}
    \caption{\textbf{Class discriminability.} Category-specific visualizations of all methods at a fixed 10\% of important pixels insertion. The original image contains exactly two categories -  Collie and Ibizan hound. The third category - Cockatoo - is the minimal activated category.}
    \vspace{-10px}
    \label{fig:class-specific}
\end{figure}

A reasonable explanation method should be able to produce discriminative visualizations for different class of interest~\citep{Grad-CAM}.
Figure~\ref{fig:class-specific} shows a category-specific visual comparison of all methods with 10\% of insertion on an image with two classes: collie and Ibizan.
We also display the visual explanation with a minimal activated class: cockatoo.
Clearly, Gradient, DeepLIFT, IG, and XRAI hardly generate class-specific explanations, since the significant pixels only slightly change when different labels are assigned.
A-FMI is able to output class discriminative explanations, which evidently shows that the relationships between input and predictions are successfully captured.
Furthermore, we find that LRP is prone to outlining the edges even for background (\cf Figure~\ref{fig:introduction}), while Grad-CAM and A-FMI output tightly identify a region of interest in the image. 

\begin{figure}
    \centering
    \includegraphics[width=\linewidth]{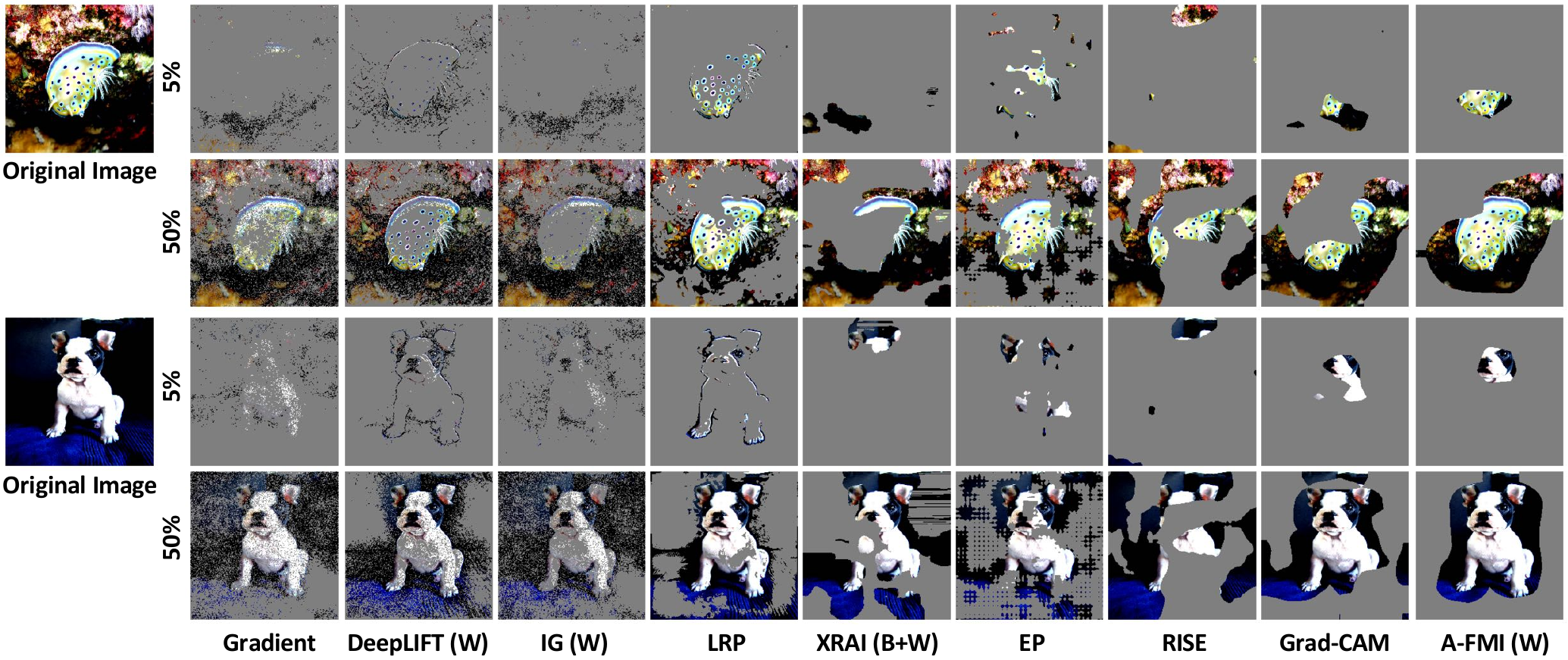}
    \caption{\textbf{Visual Inspections.} A visual comparison of all methods, where 5\% and 50\% important pixels are included, respectively. See more examples in Appendix D.
    }
    \vspace{-15px}
    \label{fig:overall-performance}
\end{figure}

\subsection{Overall Performance Comparison (RQ4)}\label{sec:overall-performance}

\textbf{Visual Inspections}. Figure~\ref{fig:overall-performance} shows that:
1) For pixel-wise attribution methods, at 5\% insertion, DeepLIFT and LRP initially capture outlines, while Gradient and IG fail to find any meaningful patterns.
As the percentage of pixel insertion increases, Gradient, DeepLIFT, and IG become more defined, however, the salient pixels are largely distributed to unrelated areas; meanwhile, LRP tends to overvalue the lines, edges and corners of the images.
2) For region-level attribution methods, XRAI, EP and RISE output more compact regions. However, thier performances are unstable.
3) For feature map-based methods, Grad-CAM and A-FMI tend to initially search for the significant characteristics of target object first (eyes or nose for dog) and then expand outward (body for dog).
Moreover, the explanations of A-FMI are more targeted than Grad-CAM, which suffers from the gradient saturation problem. This verifies the rationality and effectiveness of using references.
Sufficient visualizations of all methods with thresholds that vary from 5\% to 50\% are provided in Appendix D.


\begin{wrapfigure}{r}{0.5\textwidth}
    \vspace{-15pt}
    \begin{center}
        \includegraphics[width=0.235\textwidth]{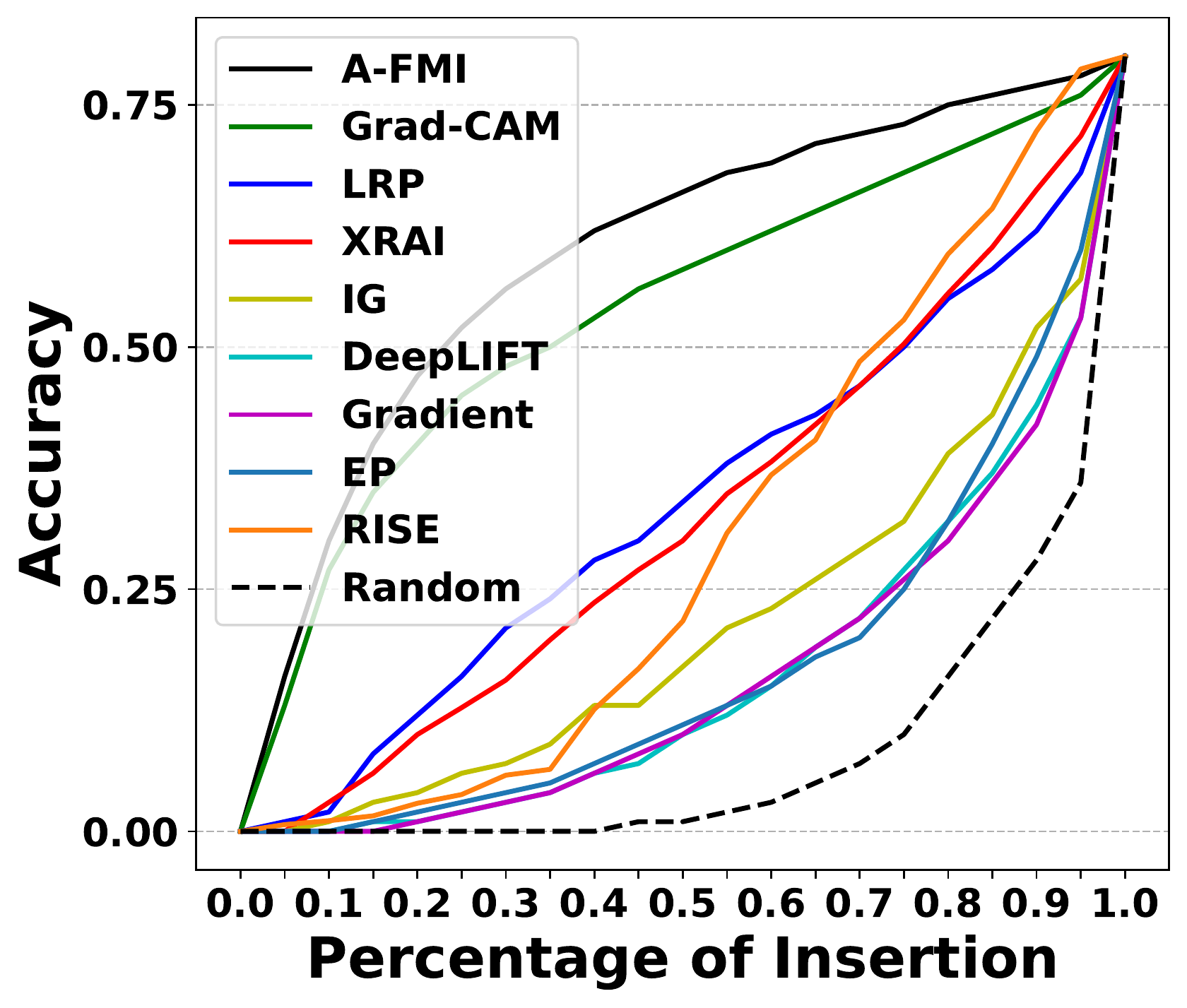}
        \includegraphics[width=0.235\textwidth]{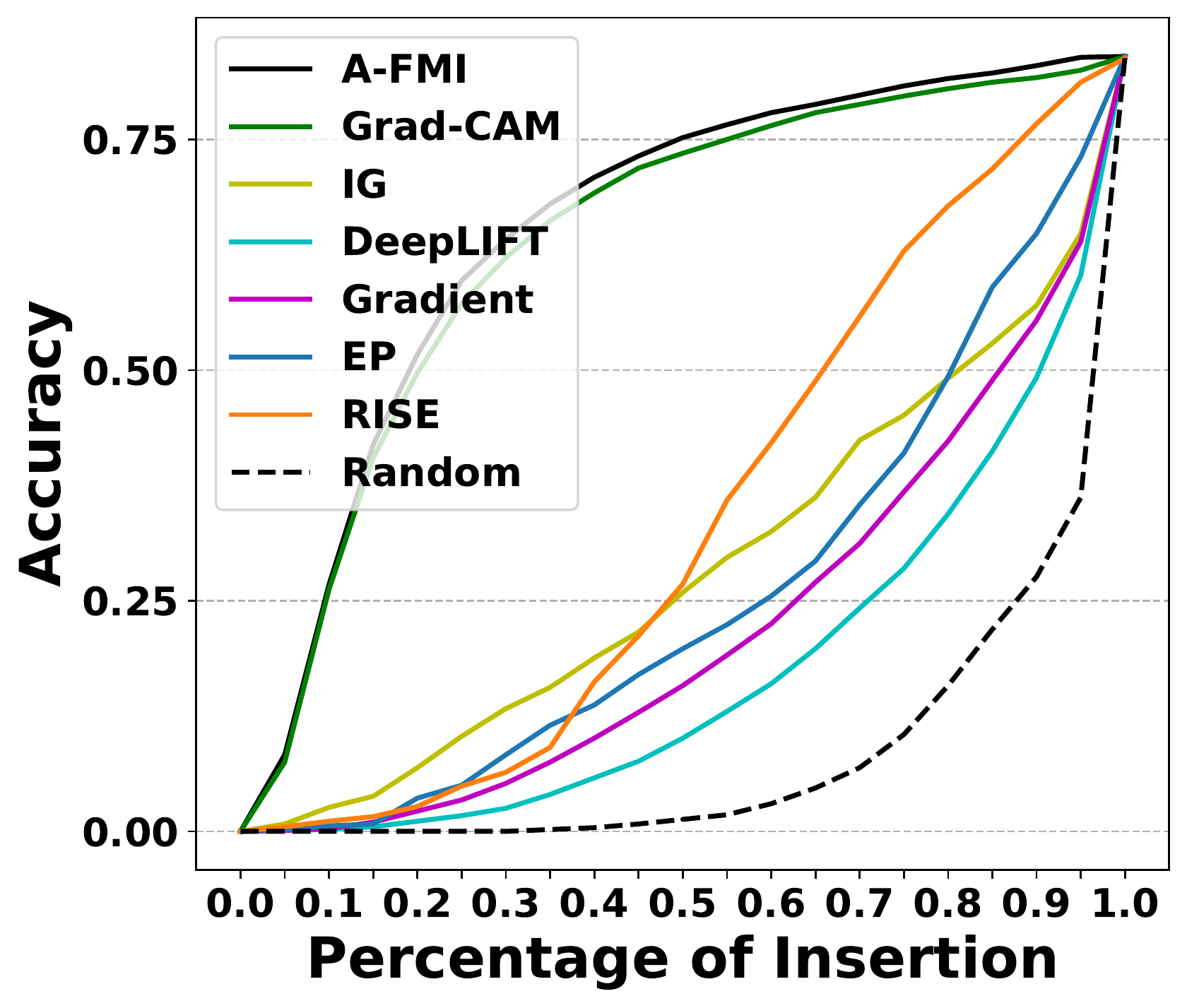}
    \end{center}
    \vspace{-15pt}
    \caption{Accuracy curves of various attribution methods \wrt different percentage of important pixel insertion (Left: VGG19; Right: ResNet50).}
    \label{fig:acc-esr}
    \vspace{-15pt}
\end{wrapfigure}

\textbf{Predictive Performance}. \emph{Insertion} metric is used to quantitatively evaluate the explanations. \emph{Deletion} metric is not used here since \cite{deletion} has shown its drawbacks.
In particular, we start with a black image, gradually add the pixels with high confidence, and feed this masked image into the CNN model.
As the percentage of pixel insertion increases from 0\% to 100\%, we monitor the changes to classification accuracy (\ie the fraction of masked images that are correctly classified). A sharp increase as well as a higher area under the accuracy curve indicate a better explanation.
In this experiment, we focus on 5000 images sampled from ImageNet and show the results in Figure~\ref{fig:acc-esr} and Table~\ref{tab:auc}\footnote{The results are different from that reported in~\cite{RISE}, where blurred cancaves and probability curve are adopted for insertion.}.
We find that Grad-CAM and A-FMI consistently outperform the other attribution methods, being able to identify pixels that are truly important to CNN as those with the highest attributions chosen by the methods.
We attribute this success to the Strong Relevance characteristic of feature maps.
Moreover, in VGG19, A-FMI achieves signiﬁcant improvements over Grad-CAM, indicating that using FMI with the reference is a promising solution to the saturation problem; whereas, A-FMI outperforms slightly better than Grad-CAM in ResNet50. This is reasonable since one fully connected layer is involved in ResNet50 and has a minor saturation problem.
We also find that pixel-wise attribution methods perform relatively poor which is consistent to the visual inspections.

\begin{table}[t]
    \caption{Performance comparison \wrt area under curves of Accuracy.}
    \label{tab:auc}
    \resizebox{\textwidth}{!}{
    \begin{tabular}{l|ccccccccccc}
    \hline
    Accuracy-AUC & Random & Gradient & DeepLIFT & IG & LRP & XRAI & EP & RISE & GradCAM & \textbf{A-FMI} \\ \hline\hline
    VGG19 & 0.0854 & 0.1657 & 0.1673 & 0.2177 & 0.3460 & 0.3269 & 0.1775 & 0.2596 &  0.5343 & $\mathbf{0.5908}^{*}$ \\ 
    ResNet50 & 0.0854 & 0.2236 & 0.1812 & 0.2857 & - & - & 0.2613 & 0.2970 &  0.6380 & $\mathbf{0.6513}^{*}$ \\ \hline\hline
    Time per image/s & - & 0.0354 & 0.0669 & 1.5908 & 0.9719 & 14.6469 & 14.1818 & 10.8123 & 0.0291 & 0.1392 \\ \hline
    \end{tabular}}
    \vspace{-10px}
\end{table}

\textbf{Computational Performance}.
In terms of time complexity, we report the time cost per image of each attribution method and find that, A-FMI performs similarly to the pixel-wise attribution methods while being significantly faster than the region-level attribution methods.
\section{Conclusion}
We proposed a novel attribution method via feature map importance, A-FMI, to produce visual explanations for CNNs.
A-FMI provides a potential way to solve the gradient saturation problem at the granularity of feature maps, which allows the information to be backpropagated even when the gradient approaches zero.
Extensive experiments illustrated the superior performance of A-FMI from both interpretable and faithful perspectives, compared with other popular attribution methods.
Future work includes
applying the attribution methods to other types of neural networks, such as graph neural networks.

\bibliography{ms}
\bibliographystyle{iclr2021_conference}

\newpage
\appendix
\section*{APPENDIX}
In this appendix, we first report the reproducibility notes of A-FMI, including the implementation details, baselines, and evaluation metrics.
In what follows, we present the visual inspections of A-FMI on different CNN Models to show the generalization ability.
We then present more comparisons between A-FMI and attribution methods \wrt reference reliability and overall performance.

\subsection*{A\quad Reproducibility}
\subsection*{A.1 Implementation Details}
\textbf{Running Environment}.
The experiments are conducted on a single Linux server with 80 Intel(R) Xeon(R) CPU E5-2698 v4 @ 2.20GHz and use a single NVIDIA Tesla V100-SXM2 32GB.
Our proposed A-FMI is implemented in Python 3.7 and Pytorch 1.2.0, and we have provided the codes in the supplementary material.

\textbf{Datasets}.
We use two datasets in Torchvision to validate the effectiveness of attribution methods: MNIST~\citep{MNIST} and ImageNet~\citep{ImageNet}, in the context of image classification. We remain the training and validation sets as the original.

\textbf{CNN models of Interest}.
On the MNIST dataset, we use the CNN example provided in the Pytorch tutorial\footnote{\url{https://github.com/pytorch/examples/tree/master/mnist}.} as the model being interpreted.
On the ImageNet dataset, we use the pretrained CNNs in Torchvision\footnote{\url{https://pytorch.org/docs/stable/torchvision/index.html}.} as the models being interpreted, which include VGG-16~\citep{vgg-19}, VGG-19~\citep{vgg-19}, ResNet-50~\citep{ResNet}, ResNet-152~\citep{ResNet}, and ResNeXt-101-32x8d~\citep{ResNeXt-101-32x8d}. These CNNs are fixed during the interpreting process.

\subsection*{A.2 Baselines}
We give detailed attribution methods as follows and include the implementations in the supplementary material. For each method, the hyperparameters are set as the original papers suggested.
\begin{itemize}
    \item \textbf{Gradient}~\citep{gradients}. This method computes the gradient of the target class \wrt each pixel as its attribution score. We use the codes\footnote{\url{https://github.com/jacobgil/pytorch-grad-cam/blob/master/gradcam.py}.} released by the authors of Grad-CAM.

    \item \textbf{DeepLIFT}~\citep{DeepLIFT}. This backpropagation-based method assigns an attribution score to each unit, to reflect its relative importance at the original neural network input to the reference input. We use the codes released in the SHAP library\footnote{\url{https://github.com/slundberg/shap}.}, where a black image is set as the default reference and the ``rescale rule'' is used to calculate the attribution.

    \item \textbf{IG}~\citep{IG}. Integrated Gradient (IG) computes the modified gradient of the class \wrt each input pixel as the attribution score. More specifically, the modified gradient is the average gradient while the input varies along a linear path from a reference input to the original input. We use the codes\footnote{\url{https://github.com/TianhongDai/integrated-gradient-pytorch}.}, where a black image is as the reference and the path steps are 100.

    \item \textbf{LRP}~\citep{LRP}. Layer Relevance Propagation (LRP) starts at the output layer, redistributes the prediction of interest as the relevance of units in the previous layer, and recursively propagates the relevance scores until the input layer is reached. We use the official implementation \footnote{\url{http://heatmapping.org/tutorial/}.}, where the ``$\epsilon$-rule'' is adopted to establish the relevance of single pixels. 

    \item \textbf{XRAI}~\citep{XRAI}. This method treats the gradients of the target prediction \wrt region-level features as the attribution scores. In particular, it coalesces smaller regions into larger segments based on the maximum gain of IG per region. We use the official implementation\footnote{\url{https://github.com/PAIR-code/saliency}.}.

    \item \textbf{EP}~\citep{EP}. Extremal Perturbations (EP) optimize a spatial perturbation mask with a fixed area and smooth boundary that maximally affects a CNN's prediction. We use the official implementation\footnote{\url{https://github.com/facebookresearch/TorchRay}.}.

    \item \textbf{RISE}~\citep{RISE}. Randomized Input Sampling for Explanation (RISE) masks an image using random occlusions patterns and observes change of confidence scores. For masking generation, RISE first samples smaller binary masks and then upsample them to larger resolution using bilinear interpolation. We use the official implementation\footnote{\url{https://github.com/eclique/RISE}.}.
    
    \item \textbf{Grad-CAM}~\citep{Grad-CAM}. This method uses the gradient of the prediction \wrt feature maps in the last convolutional layer as the attribution scores, and arranges the attribution to the input image as the saliency map. We use the codes\footnote{\url{https://github.com/jacobgil/pytorch-grad-cam/blob/master/gradcam.py}.} released by the authors.
\end{itemize}

\subsection*{A.3 Evaluation Metrics}
Going beyond the visual inspections, we qualitatively evaluate the explanations of attribution methods on the task of image classification. Specifically, two evaluation protocols, Classification Accuracy@PI and Softmax Ratio@PI, can be computed as follows:
\begin{enumerate}
    \item We start with a black image $\bar{\Mat{x}}$;
    \item For each image $\Mat{x}$ with the ground truth label $c$ in the validation set $\Set{V}$, we form a ranking list of pixels $\Mat{m}$ in descending order, based on the saliency map of an attribution method $\Mat{S}^{c}(\Mat{x})$;
    \item We add top PI (percentage of insertion) significant pixels $\Mat{m}_{\text{PI}}$ to the black image and get a masked image $\Mat{x}\odot\Mat{m}_{\text{PI}}$, where $\odot$ is the element-wise product;
    \item We then feed the masked image into the CNN model and get the prediction $\tilde{c}$ and distribution over the classes $\tilde{\Mat{p}}$.
    \item We calculate two protocols as:
    \begin{gather}
        \textbf{Classification Accuracy@PI}=\frac{1}{|\Set{V}|}\sum_{\Mat{x}\in\Set{V}}\Space{I}(\tilde{c}=c),
    \end{gather}
    where $\Space{I}(\tilde{c}=c)$ is the binary indicator to evaluate whether the masked image is accurately classified, in order to measure the quality of explanations at a coarse granularity; and
    \begin{gather}
        \textbf{Softmax Ratio@PI}=\frac{1}{|\Set{V}|}\sum_{\Mat{x}\in\Set{V}}\frac{\tilde{p}_{c}}{p_{c}},
    \end{gather}
    where $\tilde{p}_{c}$ and $p_{c}$ are the probability that the classes of masked or original images are equal to the ground truth $c$, in order to measure the attribution performance at a finer granularity.
\end{enumerate}

In our experiment, we set PI as $\{5\%,10\%,15\%,\cdots,90\%,95\%,100\%\}$ and monitor the changes of each attribution method \wrt the protocols.
Thereafter, we calculate the area under curves (AUC) of these two protocols, termed \textbf{Accuracy-AUC} and \textbf{Softmax-AUC} respectively.

\subsection*{B\quad Visual Inspections of A-FMI on Different CNN Models}
We analyze the generalization ability of A-FMI by interpreting different CNN architectures, including the VGG-16, VGG-19, ResNet-50, ResNet-152, and ResNeXt-101-32x8d models.

\begin{figure}[th]
    \centering
    \includegraphics[width=\linewidth]{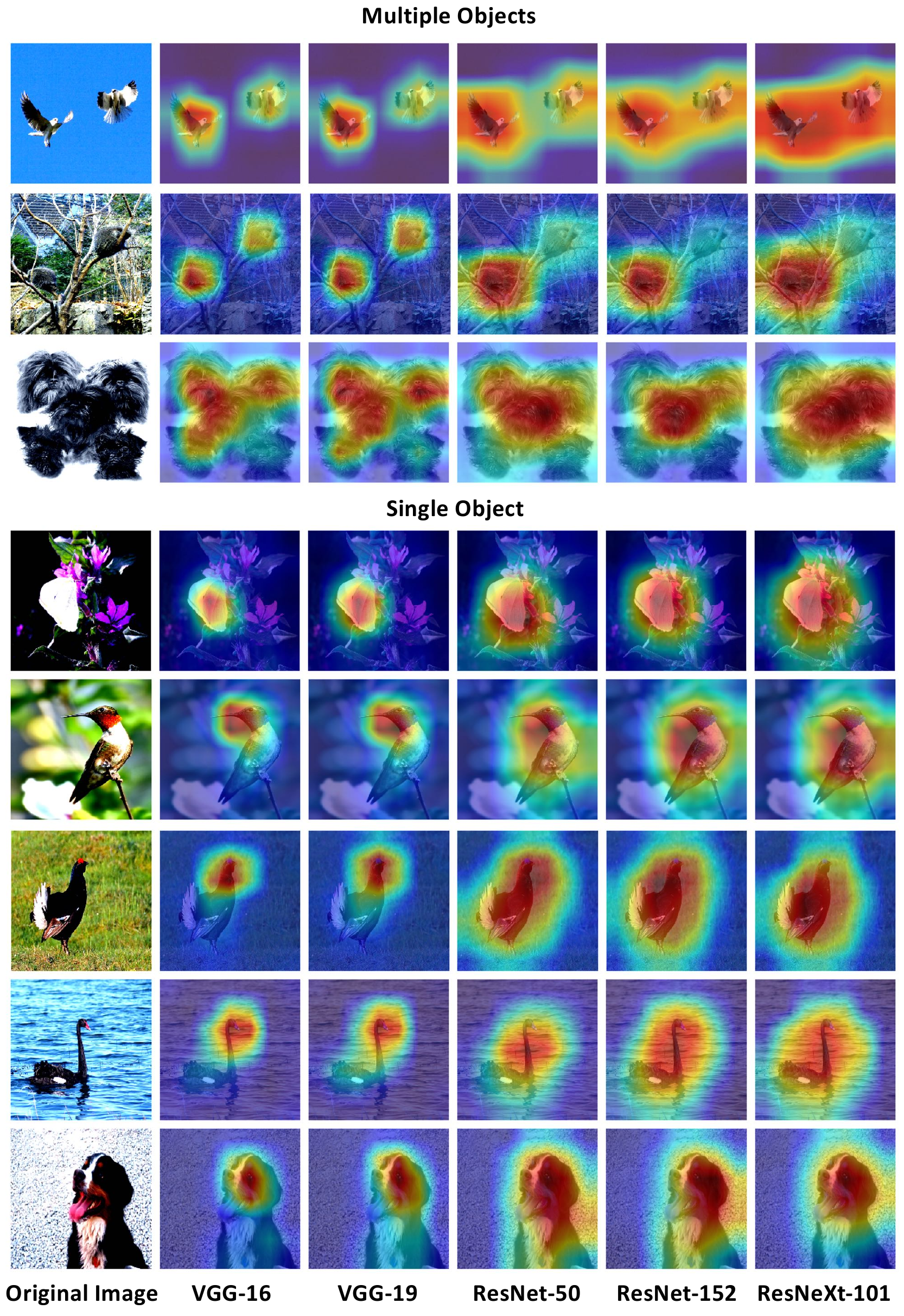}
    \caption{Visual Inspections of A-FMI on the VGG-16, VGG-19, ResNet-50, ResNet-152, and ResNeXt-101-32x8d models.} 
    \label{fig:different_cnns}
\end{figure}

Figure \ref{fig:different_cnns} shows the results of these five CNN models.
When multiple objects of interest appear in a single image, VGG-16 and VGG-19 tend to distinguish them separately and perform better localization;
meanwhile, ResNet-50, ResNet-152 and ResNeXt-101-32x8d produce a larger region to cover several objects.
When it comes to the single-object image, VGG-16 and VGG-19 might focus on the significant characteristics of the objects (\eg heads of animals), while ResNet-50, ResNet-152, and ResNeXt-101-32x8d favor finding the whole body of the object.

\subsection*{C\quad Reference Reliability}

We consider the variants of A-FMI, DeepLIFT, and IG that use different references --- a black image, a white image, and an image filled with random pixels to generate the saliency maps. The results in Figure~\ref{fig:reference} correspond to the VGG-19 network. 

Regardless of references, we observe that DeepLIFT and IG create grainy saliency maps and focus more attributions on the background or irrelevant objects (\eg the white snow background in the 1st column; the branch and grass highlighted in the 2nd column; the black background in the 3rd column; the person in the 4th column;) than within the objects of interest.

\begin{figure}[t]
    \centering
    \includegraphics[width=0.77\linewidth]{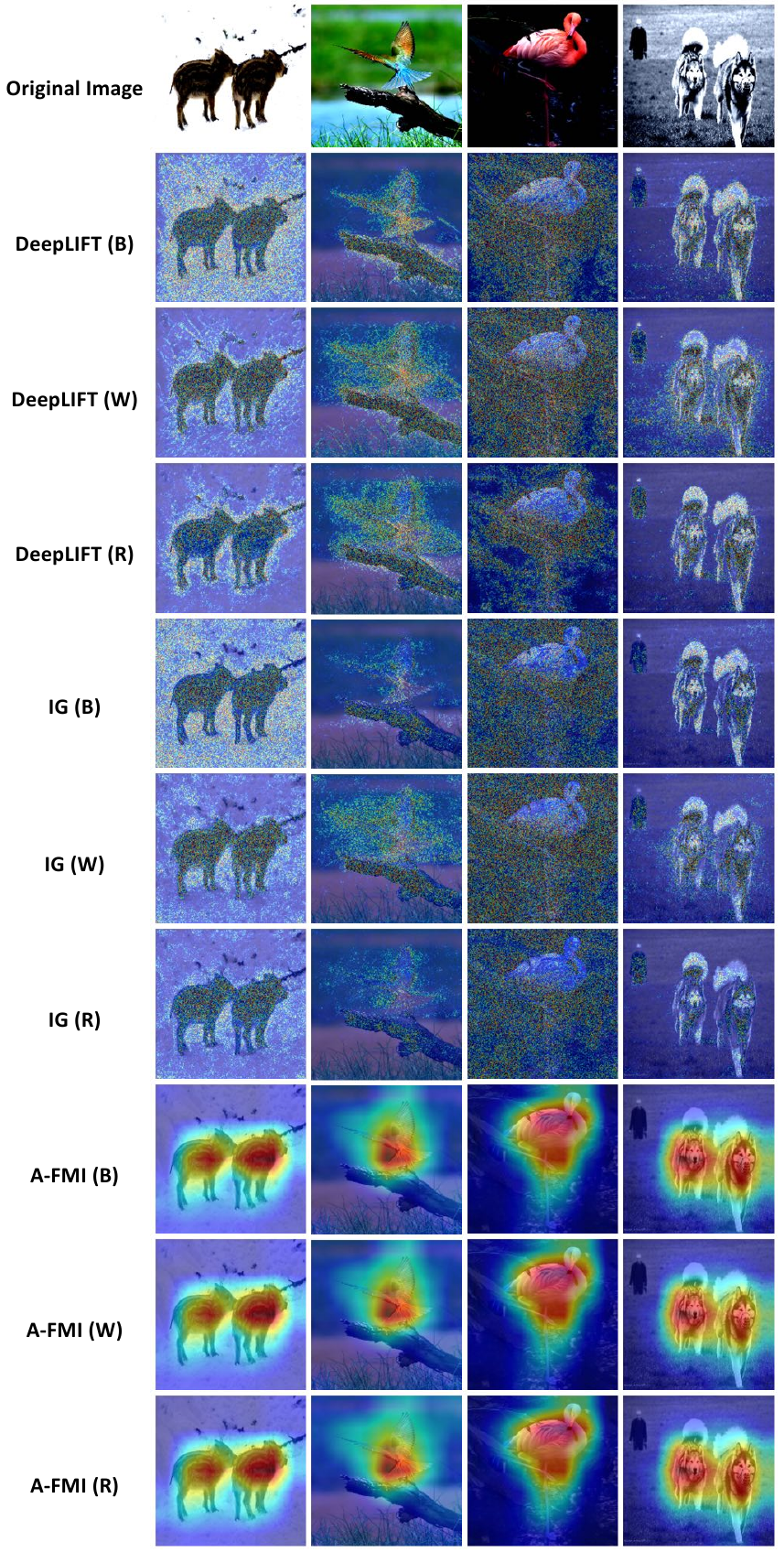}
    \caption{\textbf{Reference reliability.} A-FMI is reliable for the different choices of reference, while the explanations of DeepLIFT and IG rely heavily on the choice of references. Wherein, B, W, and R stand for black, white, and random reference respectively.} 
    \label{fig:reference}
\end{figure}

\subsection*{D\quad Overall Performance}
\begin{figure}[t]
    \centering
    \includegraphics[width=\linewidth]{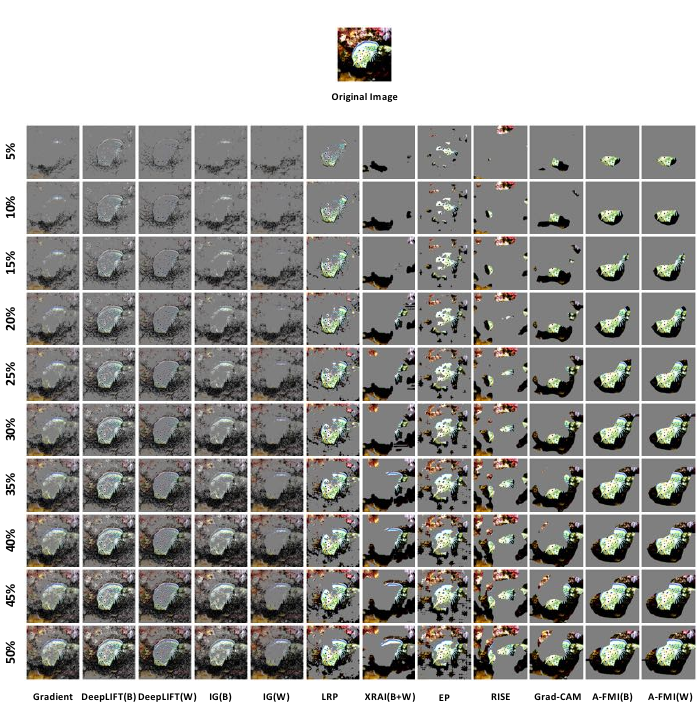}
    \caption{\textbf{Visual Inspections.} A visual comparison of all methods, where the percentage of important pixel insertion varies from 5\% to 50\%. Gradient, DeepLIFT and IG tend to produce grainy images. LRP, XRAI and Grad-CAM might choose disconnected areas. A-FMI focuses more on the objects of interest.
    } 
    \label{fig:appendix-1}
\end{figure}

\begin{figure}[t]
    \centering
    \includegraphics[width=\linewidth]{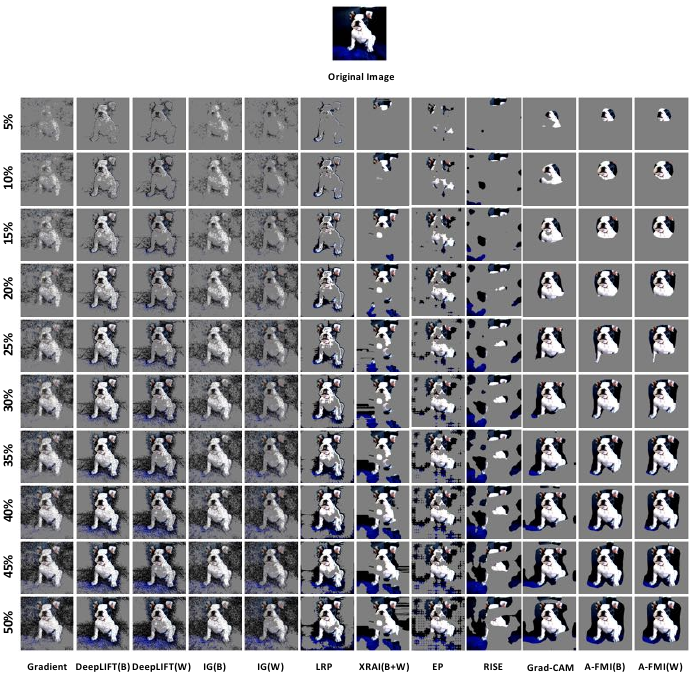}
    \caption{\textbf{Visual Inspections.} A visual comparison of all methods, where the percentage of important pixel insertion varies from 5\% to 50\%. Gradient, DeepLIFT and IG tend to produce grainy images. LRP, XRAI and Grad-CAM might choose disconnected areas. A-FMI focuses more on the objects of interest.} 
    \label{fig:appendix-2}
\end{figure}

\begin{figure}[t]
    \centering
    \includegraphics[width=\linewidth]{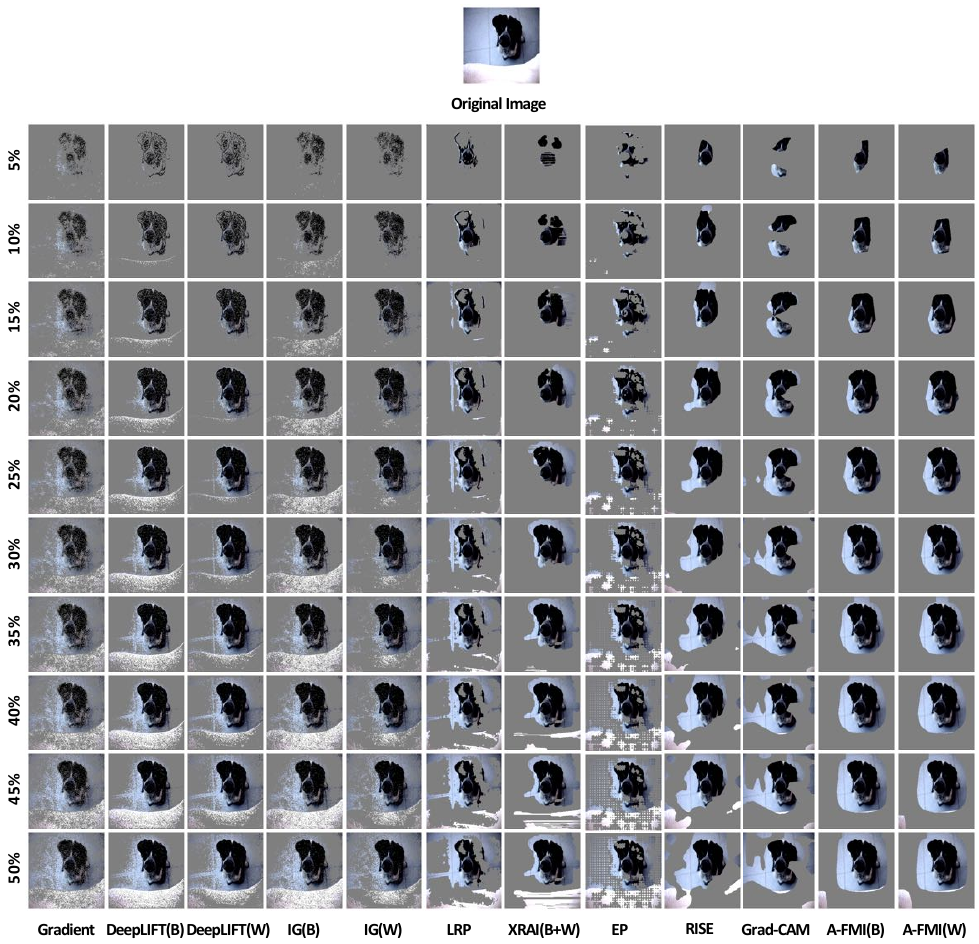}
    \caption{\textbf{Visual Inspections.} A visual comparison of all methods, where the percentage of important pixel insertion varies from 5\% to 50\%. Gradient, DeepLIFT and IG tend to produce grainy images. LRP, XRAI and Grad-CAM might choose disconnected areas. A-FMI focuses more on the objects of interest.} 
    \label{fig:appendix-3}
\end{figure}

\begin{figure}[t]
    \centering
    \includegraphics[width=\linewidth]{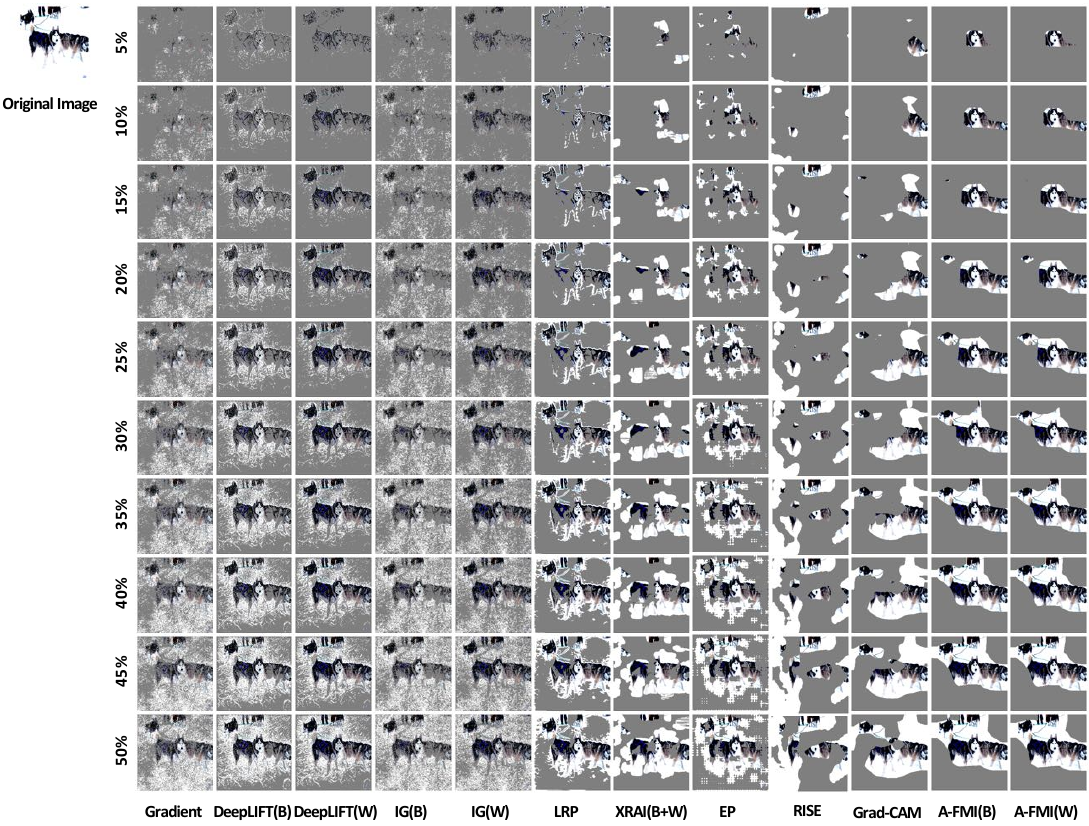}
    \caption{\textbf{Visual Inspections.} A visual comparison of all methods, where the percentage of important pixel insertion varies from 5\% to 50\%. Gradient, DeepLIFT and IG tend to produce grainy images. LRP, XRAI and Grad-CAM might choose disconnected areas. A-FMI focuses more on the objects of interest.} 
    \label{fig:appendix-4}
\end{figure}

\begin{figure}[t]
    \centering
    \includegraphics[width=\linewidth]{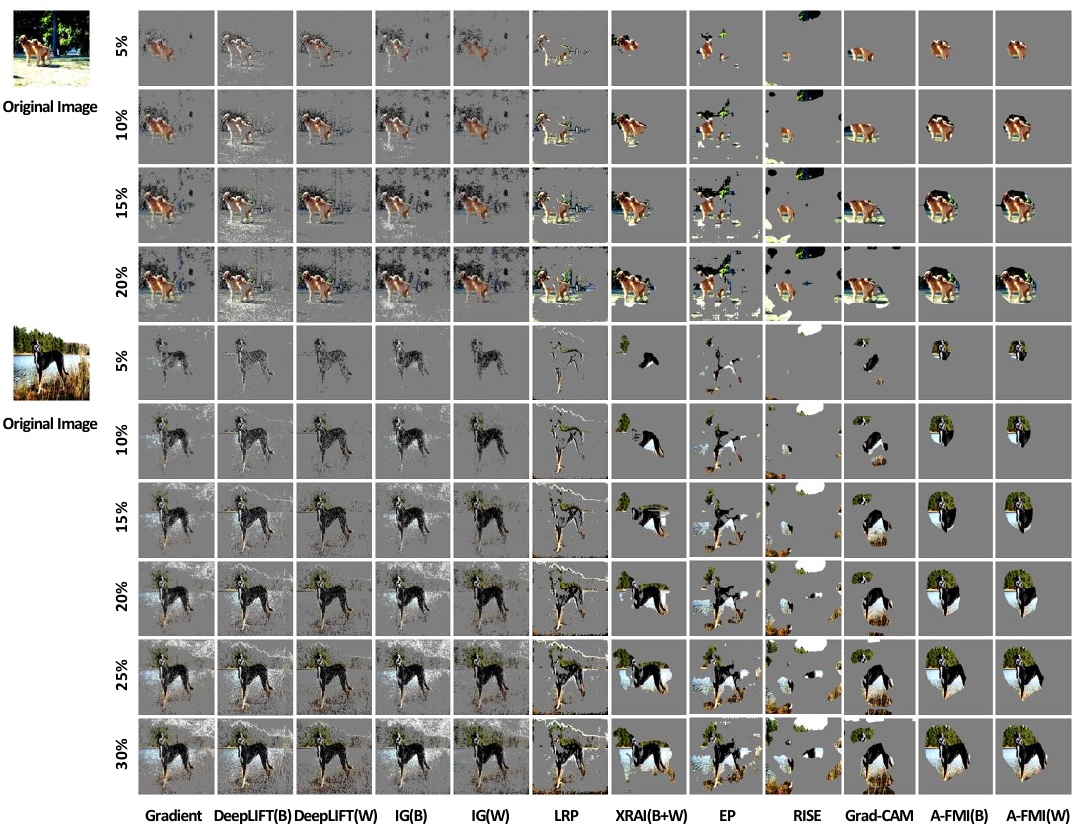}
    \caption{\textbf{Visual Inspections.} A visual comparison of all methods, where the percentage of important pixel insertion varies from 5\% to 50\%. Gradient, DeepLIFT and IG tend to produce grainy images. LRP, XRAI and Grad-CAM might choose disconnected areas. A-FMI focuses more on the objects of interest.} 
    \label{fig:appendix-5}
\end{figure}

We show visualizations of all methods with a percentage of important pixels insertion that vary from 5\% to 50\% in VGG-19. We compare A-FMI to popular attribution methods at three granular levels: pixel-wise (Gradient, DeepLIFT, IG, LRP), region-level (XRAI, EP, RISE), and feature map-based (Grad-CAM).
Figures \ref{fig:appendix-1}-\ref{fig:appendix-5} clearly show that, compared with other popular attribution methods, A-FMI provides better visual explanations, in terms of more accurate and targeted localization of objects. 

\begin{figure}[h]
	\centering

	\subcaptionbox{Softmax Ratio curves on VGG19\label{fig:19-acc}}{
		\includegraphics[width=0.32\textwidth]{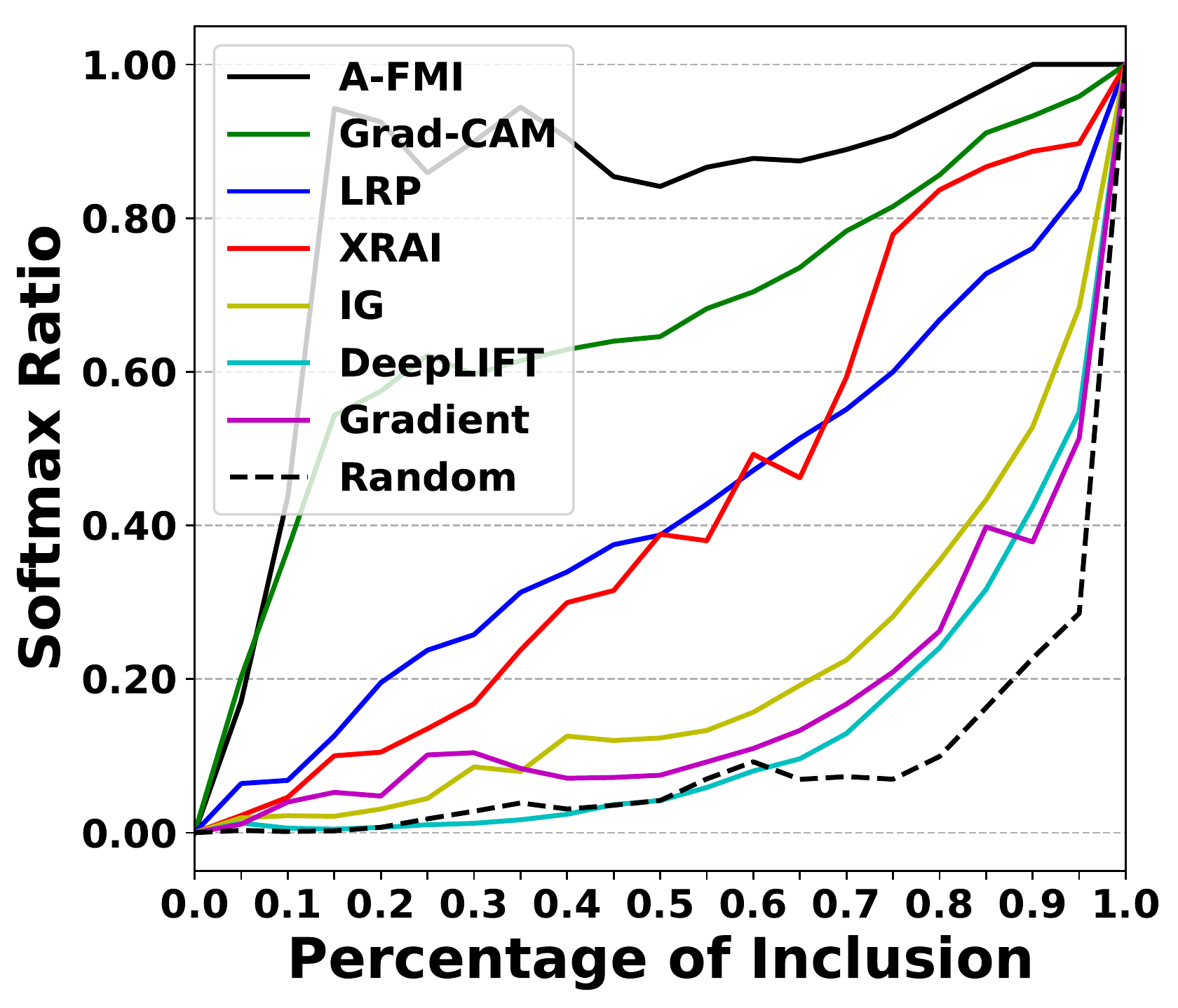}}
	\subcaptionbox{Accuracy \& Softmax Ratio curves on VGG16\label{fig:16-acc}}{
		\includegraphics[width=0.32\textwidth]{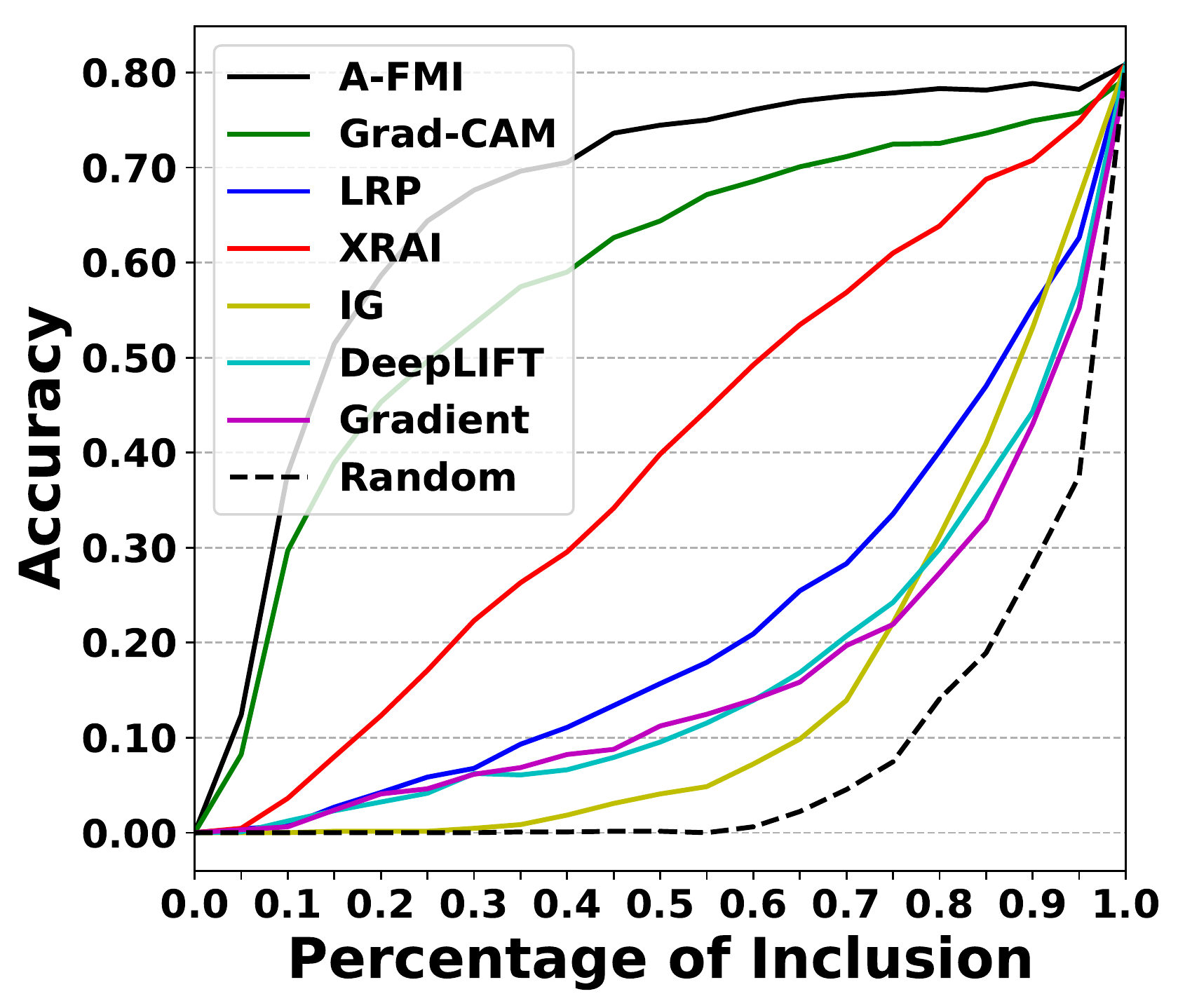}
		\includegraphics[width=0.32\textwidth]{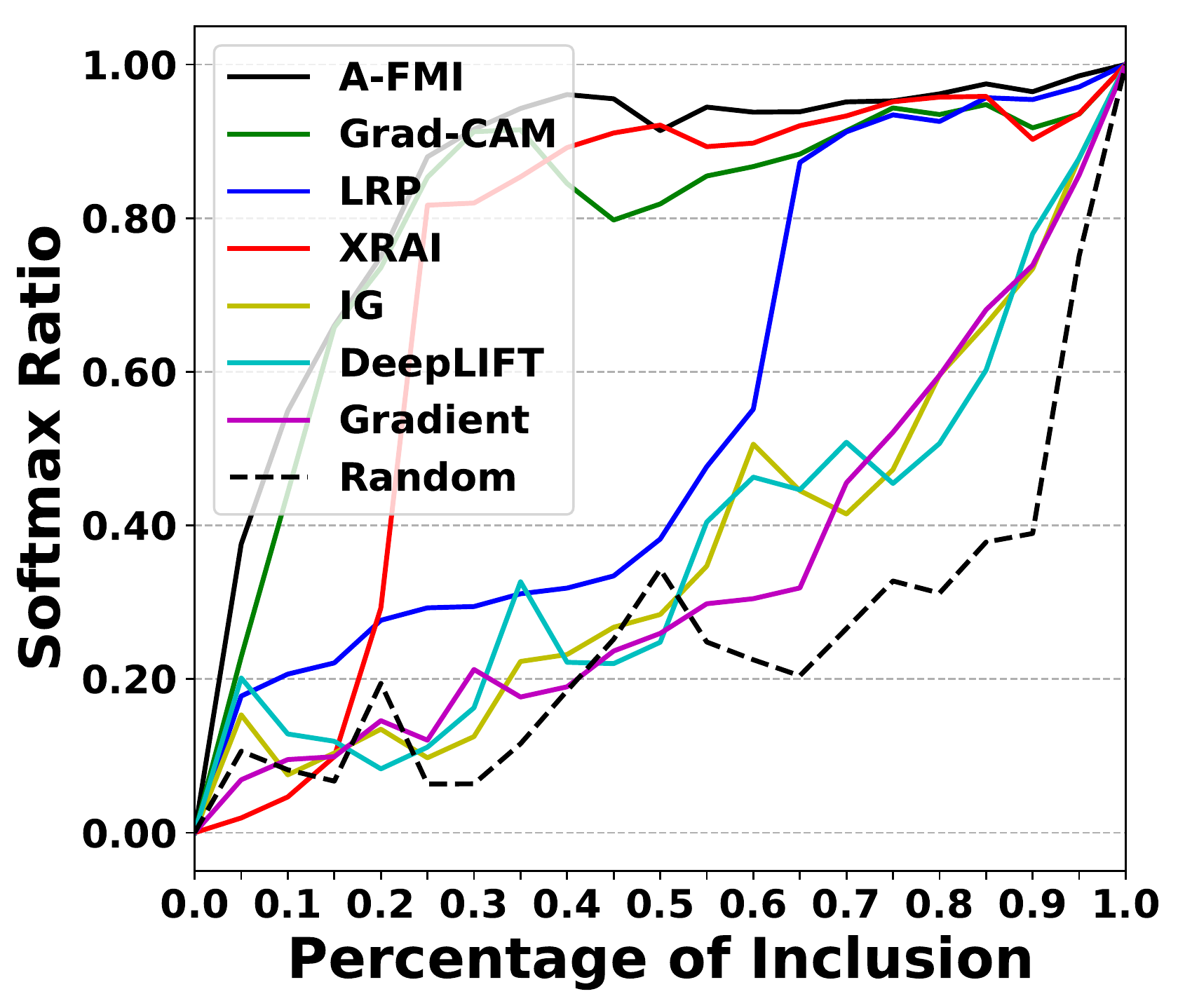}}
	\vspace{-5pt}
	\caption{\textbf{Predictive Performance}. Accuracy curves and Softmax ratio curves of various attribution methods \wrt different percentage of important pixels insertion. Best viewed in color.}
	\label{fig:acc-esr}
	\vspace{-15pt}
\end{figure}

\end{document}